# Detection and Attention: Diagnosing Pulmonary Lung Cancer from CT by Imitating Physicians

Ning Li*, Haopeng Liu*, Bin Qiu, Wei Guo, Shijun Zhao, Kungang Li †, Jie He †

*Abstract*—**This paper proposes a novel and efficient method to build a Computer-Aided Diagnoses (CAD) system for lung nodule detection based on Computed Tomography (CT). This task was treated as an Object Detection on Video (VID) problem by imitating how a radiologist reads CT scans. A lung nodule detector was trained to automatically learn nodule features from still images to detect lung nodule candidates with both high recall and accuracy. Unlike previous work which used 3-dimensional information around the nodule to reduce false positives, we propose two simple but efficient methods, Multi-slice propagation (MSP) and Motionless-guide suppression (MLGS), which analyze sequence information of CT scans to reduce false negatives and suppress false positives. We evaluated our method in open-source LUNA16 dataset which contains 888 CT scans, and obtained state-of-the-art result (Free-Response Receiver Operating Characteristic score of 0.892) with detection speed (end to end within 20 seconds per patient on a single NVidia GTX 1080) much higher than existing methods.**

## I. Introduction

About 30-40% pulmonary nodule might have connection with lung cancer. Patients who are diagnosed at an early stage of lung cancer have a significantly better survival rate than at a later stage. 5-years survival rate drops from over 80% when the lung cancer is detected in stage I to 1% when it is detected in stage IV. Hence, it is of high importance to detect the nodule in early stage.

Many endeavors have been made to use CAD system to distinguish lung nodule from other similar tissues such as blood vessel and weasand. A representative approach attempted to merge the adjacent 3 CT slices to detect lung nodule in 2D images with high recall, and reduced false positives using the 3D information around the candidate nodule [1], [2]. In another work, the author observed a nodule candidate from 9 different directions and provided diagnosis by synthesizing all the information [3]. A third method is to utilize the 3D convolution network by directly taking a 3D image of the lung as input and outputting the nodule position and probability [4]. These methods are computationally intensive even with high-end hardware (e.g., cluster of Nvidia Titan X), impeding their affordable real-world application.

In practice, radiologists observe each CT slice together with adjacent slices. Analysis of the nodule characteristics includes size, density, edges, the relationship between the surrounding tissue, and bronchial and vascular relationship. Thanks to advances in CT technology, coronal and sagittal pattern reconstruction is also applied to the clinic, providing more information to the radiologist. However, ultimately, 3-dimensional graphics need to be formed in the minds of radiologists, who then make the final judgment; this usually requires professional training and years of experience.

Because the workflow of radiologists is similar to detecting an object in video, we borrowed the main idea of VID and trained a reliable nodule detector to detect the nodule candidates on still images. Thereafter, we analyzed the correlation of detections on adjacent slices, reduced the false negatives and suppressed false positives based on the sequence information.

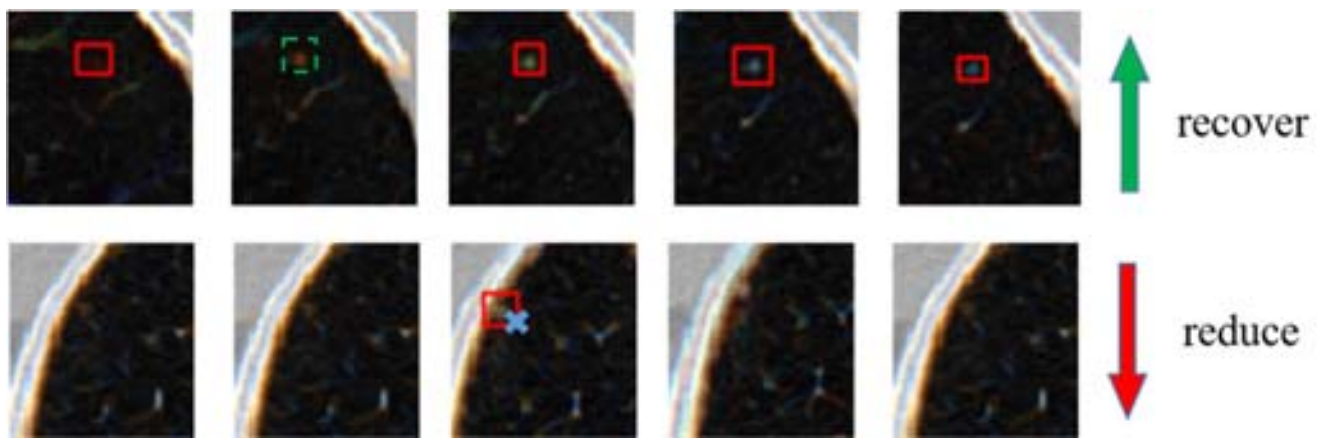

Fig.1 Top: Multi-slice propagation. Bottom: Motionless-guide suppression

Obviously, the performance of still image detector significantly influences the detection result. In recent years, the lung nodule detection algorithm has been transforming from “feature engineering” to “network engineering”. In contrast to hand-designed features (e.g., intensity thresholding, shape curvedness, and mathematical morphology [5], [6], [7]), features learned by a neural network from large-scale data (e.g. LIDC-IDRI [8]) require minimal human involvement but show state-of-the-art performance in many medical image analysis

This work is supported in part by Genonova Inc, and in part by the Department of Thoracic surgery, Cancer Hospital, Chinese Academic of Medical Science, Beijing, China.

* Ning Li and Haopeng Liu share co-first authorship

†Kungang Li and Jie He share the corresponding authorship (kungang.li@genonova.com, prof.jiehe@gmail.com)

Ning Li is with Department of Thoracic surgery, Cancer Hospital, Chinese Academic of Medical Science, 17 South of Panjiayuan, Beijing, China. (e-mail: lining@cicams.ac.cn

Haopeng Liu is with Genonova Inc (e-mail: haopeng.liu@genonova.com).

Kungang Li is with Genonova Inc (e-mail: kungang.li@genonova.com).

Jie He, Bin Qiu, Wei Guo and Shijun Zhao are with Department of Thoracic surgery, Cancer Hospital, Chinese Academic of Medical Science, Beijing, China.

applications [9], [12]. Therefore, we made efforts on designing innovative neural networks to learn rich semantic representations. We improved the state-of-the-art region-free object detection algorithm Single Shot Multibox Detector (SSD) [10] by stacking multi-level and multi-resolution features from deep convolution network (ConvNet) [11], which can overcome the inefficiency of SSD in small object detection. Experiments showed that our method was efficient to detect small lung nodules.

We proposed two simple but efficient algorithms called Multi-slice propagation (MSP) and Motionless-guide suppression (MLGS) to reduce false negatives and false positives, respectively. As shown in Fig.1, MSP recovers the missing detections by propagating the detection result to adjacent slice. MLGS builds a series of features to describe nodule motion and reduces false positives with Random Forest algorithm.

This work has three innovations: (1) We designed an efficient nodule detector on 2D image based on the SSD detector, which takes advantage of multi-level features of the ConvNet and overcomes the SSD's poor performance in detecting small object. (2) We proposed two novel algorithms, MSP and MLGS, to reduce false negatives and false positives by incorporating the temporal and context information from adjacent slices. (3) We investigated the main difference between natural and CT image processing, providing guidelines to improve the model performance for other researchers.

## II. METHOD

In this section, we firstly compare the main difference between lung nodule detection task and VID task (Section II-A). We then describe the framework of our method in Section II-B. Section II-C describes the details of our still image detector. The methods of using multi-context information and motion information to reduce false negative and false positive detections are stated in Section II-D.

### A. VID and lung nodule detection task

The ImageNet object detection from video (VID) task requires participants to localize 30 different targets in the video frames. For each video frame, the algorithm needs to produce a set of annotations $(f_i, c_i, s_i, b_i)$ of frame index $f_i$, class label $c_i$, confidence score $s_i$ and bounding box $b_i$ [11].

The lung nodule detection task is to output a set of annotations $(x_i, y_i, z_i, d_i, s_i)$ of the lung nodule center position $(x_i, y_i, z_i)$, diameter $d_i$ and confidence $s_i$. Since the task is similar to the VID task, we attempted to apply similar technologies to lung nodule detection task.

The main differences of lung nodule detection and VID task are: (1) Unlike VID task, we cannot train a reliable still image detector using only a single CT slice, because it cannot provide enough information to distinguish a nodule from other tissues. (2) Detectors designed for VID cannot be directly used for lung nodule detection, because the detector backbone (e.g., ResNet [12]) for natural images may not fit the CT images. (3) Motions exist in VID objects so that an object may be present in different locations in different frames; on the other hand, the lung nodule is motionless and is at nearly the same location in adjacent slices.

### B. Method Overview

The CAD system for lung nodule detector is shown in Fig.2. The system consists of three main parts: (1) Still image detector, (2) Multi-slice propagator, (3) Motionless-guide suppressor based on the Random Forest algorithm.

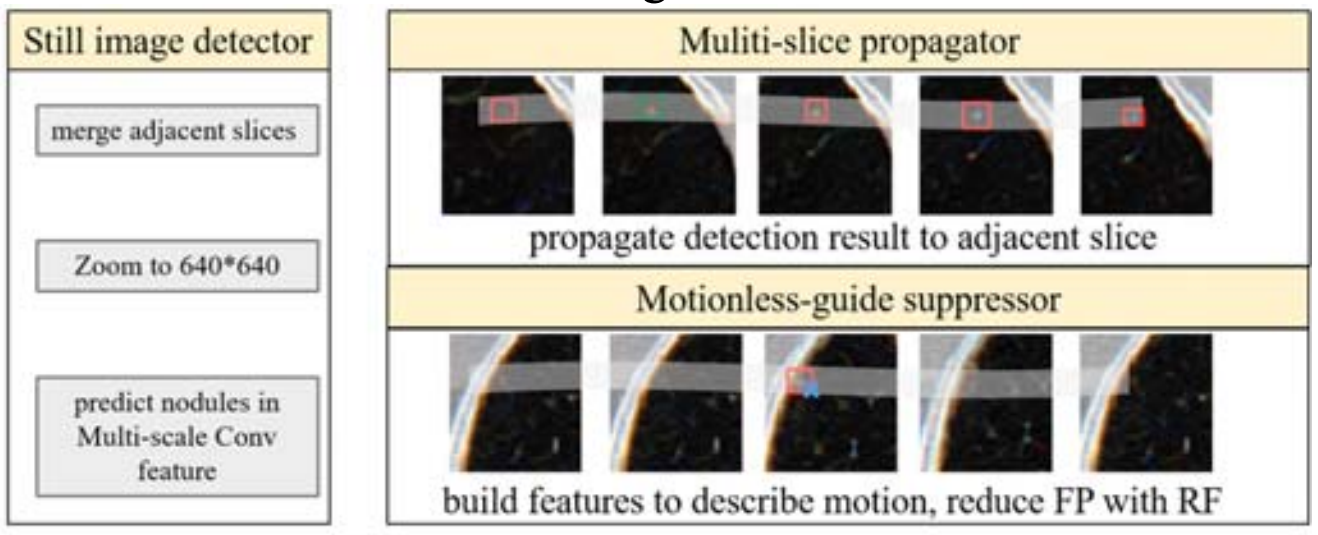

Fig.2 Workflow of our CAD system. Left: Still image detector that predicts lung nodules in multi-scale convolutional features. Right top: Multi-slice propagator that propagates detection result to adjacent slice to reduce false negative. Right bottom: Motionless-guide suppressor that builds features to describe motion, and reduces false positive with Random Forest (RF).

**Still image detector** The still image detector is based on SSD, which uses a feed-forward convolution neural network that produces fixed-size collections of bounding boxes and scores for the presence of lung nodule in those boxes, followed by a non-maximum suppression (NMS) step to produce the still image detection. We analyzed why SSD was inefficient to detect small object and found that the information of small objects was lost in high-level layer of ConvNets.

**Multi-slice propagator** In still image detection step, some nodules may be missing in certain slices but are detected in adjacent slices. Multi-slice propagator propagates the detection result to adjacent slices to recover false negatives.

**Motionless-guide suppressor** Since the lung nodule is present in nearly the same position in adjacent slices, we built motion features and used Random Forest algorithm to reduce false positives.

### C. Training still image detector from scratch

Though tremendous progress has been made in object detection, one of the remaining major open challenges is to detect small objects such as 4x4 lung nodules. The difficulty mainly results from two aspects: (1) Features. The high-level features of deep ConvNet integrate the low-level features to get rich semantic and context information, and realize high quality detection based on that information. However, small lung nodules may have little information in high-level features. In addition, low-level feature map has more information of small nodules than high-level, yet less context information to distinguish nodules from background. Also, since most image detection algorithms scan window in the feature map to generate bounding boxes and predict whether the object is present in those boxes, the high-level feature map has low resolution and it is difficult to generate sufficient bounding boxes which contain small nodules. (2) Data imbalance. Considering the lung nodule as positive samples, and other tissues as negative samples, the ratio of positive sample and negative sample is approximately 1:5000, which causes extreme data imbalance during training and make the model prediction prone to be negative. We explored some strategies to alleviate those problems when building the model.

**Detector** To our best knowledge, the two-stage object detector with region proposal (e.g., Faster RCNN [13]) is more efficient to detect small objects but slower than the one-stage detector (e.g., SSD [10]) without region proposal. However, for the lung nodule detection task, the region proposal was redundant because we only need to distinguish nodules from the background. Hence, we chose the SSD as our baseline detector, and explored how to overcome the shortcoming of SSD in detecting small objects.

**Features** We observed the importance of context information around the lung nodule. For example, the lung nodule may cause vascular traction or other lung lesion, which helps detecting the lung nodule with high confidence. It is known that the deep ConvNet aggregates low-level features to get rich semantic information in high level, but small objects barely have any information in high-level features (Fig.3 (a)). The original SSD used the pyramidal feature hierarchy to predict multi-size objects (high-level features for big object, low-level features for small object). The low-level feature contains more information of small objects but less context information (Fig.3(b)). Inspired by Feature Pyramid Network [14], we propagated high-level feature from top to bottom by Transpose Convolution [15], and integrated rich context information features and low-level features. This approach keeps the small lung nodule information while getting more context information. Another advantage of it is that the low-level feature has high resolution, which helps to generate sufficient bounding boxes that contains nodule for slide-window detector.

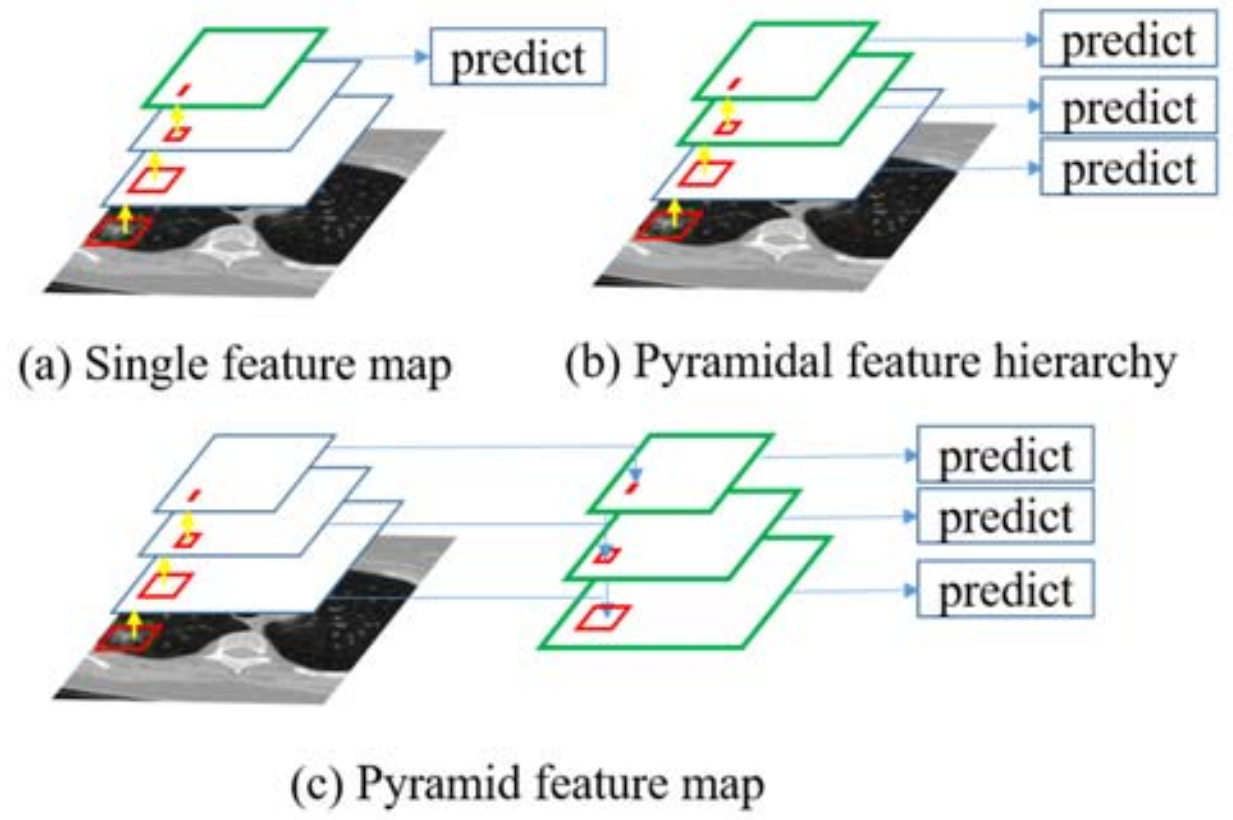

Fig.3 Top left: When using single scale high-level feature, small object has little information in it. Top right: When using the pyramidal feature hierarchy to predict multi-size objects, it is hard to get sufficient context and semantic information for small objects. Bottom: Propagating the high-level feature to low-level feature results in more context information and semantic information with high resolution.

**Data imbalance** We attempted to address the problem of data imbalance in lung nodule detection. In our experiment, we treat lung malignant nodule as positive samples, other tissues such as bone, vessels and trachea as negative samples. The positive and negative sample ratio is approximately 1:5000, which causes extreme data imbalance during training and make the model prediction prone to be negative. We explored two strategies to alleviate this problem.

**Resampling**: We resampled negative sampled during training and limited the negative to positive ratio to k:1, where k is a hyperparameter adjusted according to model performance. Tissues not selected were labeled as -1. While training, those tissues make zero contribution to the loss and will not be backpropagated.

**Loss function insensitive to data imbalance**: Since the ConvNet optimizes parameters by minimizing the loss function, it determines what information is learned by the model. The Focal Loss [16] reshapes the standard cross entropy loss such that it down-weights the loss assigned to well-classified examples. It is designed to alleviate the data imbalance problem. The Focal Loss was states as follow:

$$\mathrm{FL}(p_t) = -(1-p_t)^{\gamma}\log(p_t)$$

where $p_t$ is the probability of class t predicted by the model. The Focal Loss adds a factor $(1-p_t)^{\gamma}$ to the standard cross entropy criterion, causing misclassified examples to produce greater loss than the well-classified examples.

It is worth to note that even with the Focal Loss, if there were too many negative samples compared with positive samples, the negative samples will produce greater loss than positive samples. Therefore, we used the resampling strategy together with focal loss. As shown in Fig.4, Focal Loss enables the model to learn more negative samples in one shot, and makes it focus on hard-to-classify positive examples.

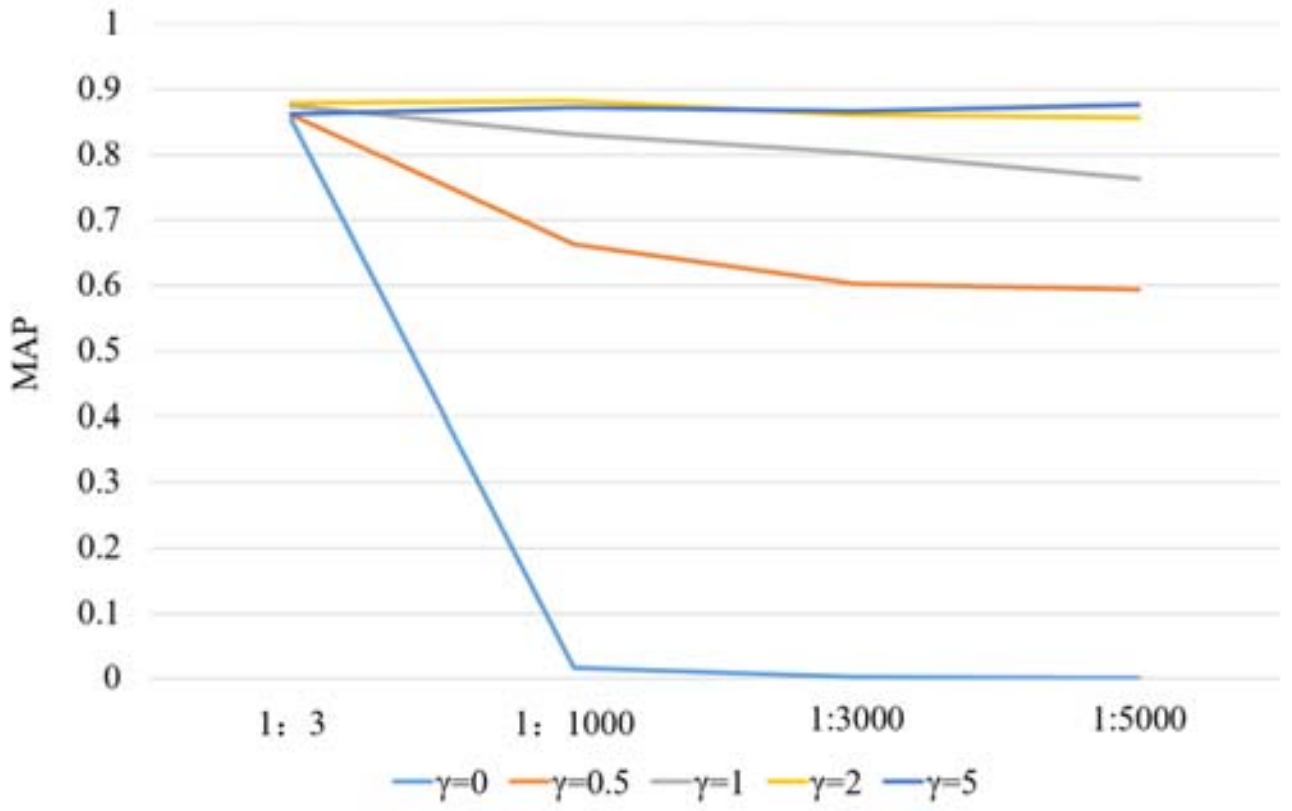

Fig.4 Model performance with different resampling ratios and Focal Loss parameters. When $\gamma = 0$, the Focal Loss degrades to the standard cross entropy.

## D. *MSP and MLGP*

**Multi-slice propagator** False negatives are typically caused by two reasons: (1) There are no bounding box that covers the small nodule. (2) Due to the small size of nodule, the detection score is low.

The false negative can be corrected by adding detection information from adjacent slices; because adjacent slices are highly correlated, the detection result should have similar spatial location. This inspires us to propagate boxes and scores of each slice to adjacent slice to augment detection and reduce false negatives. We have a false negative detection frame; the MSP with window of $w$ will propagate the detecting result of $w$ adjacent frame to the current frame. Since the lung nodule diameter is relatively small, building a short-range MSP (e.g., $w$=2) generally reduces the false negative error efficiently.

**Motionless-guide suppressor** Generally, the false positives are caused by other tissues in the lung that have similar model responses with lung nodule, such as blood vessel and weasand. The main difference between lung nodule and the false positive is that the lung nodule is motionless in adjacent slices. For example, in Fig.1, the spatial nodule location keeps changing in adjacent slices, which thus has a high probability not to be a true lung nodule; namely, it is a false positive.

TABLE I
FEATURES USED BY MLGS TO REDUCE FALSE POSITIVES

| Feature | Description |
|---|---|
| $d_{max}/n$ | $d_{max}$ is the max nodule diameter detected in 2D image; $n$ is the number of slices in which the nodule is present. Since the lung nodule is nearly spherical, $d_{max}/n$ of it should be approximately 1. |
| $p$ | The probability predicted by the still image detector |
| $d_{max}, d_{min}$ | Max and min of nodule diameters. Since the lung nodule is spherical, the $d_{max}$ should be close to $d_{min}$ |
| eccentricity ratio | The lung nodule should be present in similar locations in adjacent slices; the eccentricity ratio describes the nodule position deviation in adjacent slices. |

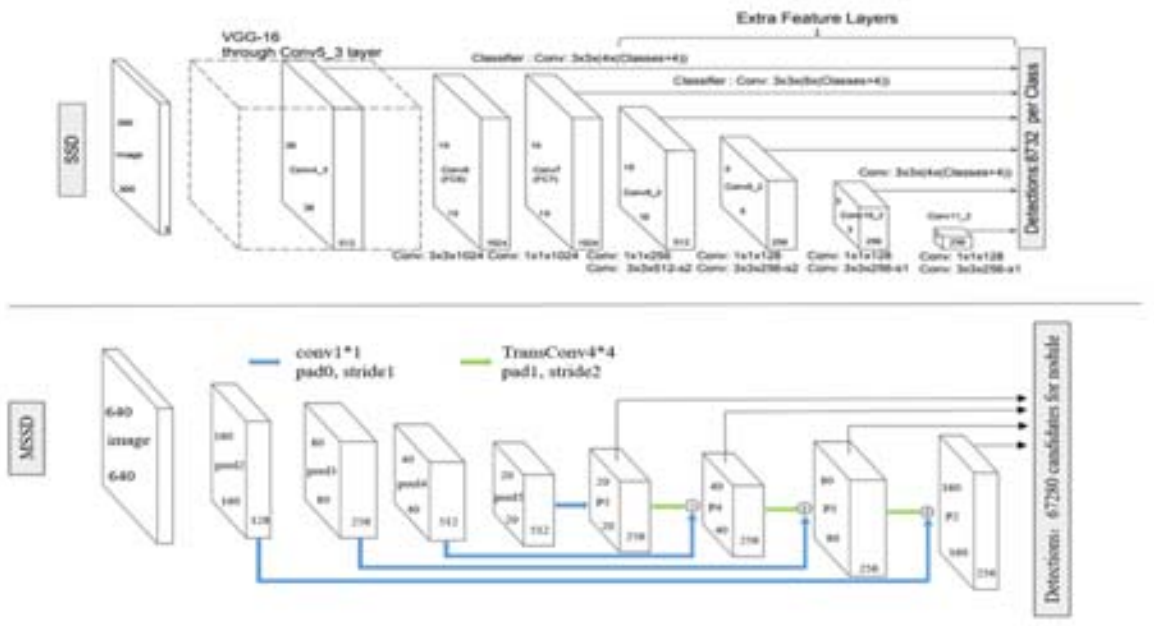

Fig.5 Our model structure with name MSSD. Compared to SSD, we propagate high-level features from top to bottom to get features with both rich context information and semantic information.

We first clustered the 2D image detection results and organized them into 3D cubes, then we built a model to describe the motion information. Key features in the model are listed in TABLE I.

We used those features to train a binary classifier, based on Random Forest algorithm, to differentiate lung nodules from normal tissues.

## III. Experiment

In this section, we firstly introduce the dataset that we use to evaluate our method (Section III-A), then we describe the still image detector experiment result in Section III-B; the MSP and MLGS experiment results are introduced in Section III-C.

### A. Dataset

We use the LUNA16 ((http://luna.grand-challenge.org/) dataset to train and evaluate our method, which is a subset of the largest public dataset for lung nodule diagnosis, the LIDC-IDRI dataset. LUNA16 removes the CT which has the slice thickness greater than 3mm, slice spacing is inconsistent or has missing slices from LIDC-IDRI. Luna16 dataset contains 888 patient CTs.

To make the CT image fit the general machine learning or deep learning algorithms, we first clip raw data to [-1200, 600], rescale the range linearly to (0, 1), and multiply it by 255. Finally, we store processed images with uint8.

### B. Still image detector

Our final model is present in Fig.5. We jointly train a binary predictor and regression detector to obtain the probability of a nodule present at image position ($x$, $y$) and its diameter. Our backbone model is based on VGG16 [17], a light-weighted version of ConvNet which has high speed for training and inference. We propagate high level features to low level features to get features with high resolution and more context information. To overcome the data imbalance problem, we use the focal loss function.

**Training** Given a CT image with ground-truth annotations, we define positive bounding boxes to be Intersection-of-Union (IOU) with ground-truth exceeding 0.5, and others to be negative. Since we use multi-scale images to detect multi-size lung nodules, it is natural to use high resolution features to detect small nodule, and low resolution features to detect big nodule. We use stochastic gradient descent method to conduct forward propagation, and compare the detection result with true labels and compute the loss, which is then backpropagated to update the model parameters. All model parameters are initialized randomly with Xavier. We train 100 epochs on the 7285 2D CT slices with initial learning rate of 0.001, momentum of 0.9 and weight decay of 0.0005. After 75 epochs, we change the learning rate to 0.001. It takes about 10 hours to reach model convergence with 4 GTX Titan X.

**Key point to improve model performance** The CT image is different from the natural image. Whether the deep learning algorithm designed for natural image also works for CT images is an open question. We tested a variety of deep learning algorithms on CT images and checked their performances. We proposed empirical explanation, which is expected to help other researchers to improve their model performance.

**Batch Normalization and Dropout** With Batch Normalization [18], normalization is performed in each training mini-batch. It allows us to use much higher learning rate and makes the model insensitive to initialization parameters. The dropout method [19] randomly drops units from the neural network during training and prevents the neural network from overfitting. We present the training performance of various approaches in Fig.5. The Batch Normalization significantly accelerates the training (convergence in 40 epochs), and

Dropout makes the model performance better in longer training time (convergence in 80 epochs)

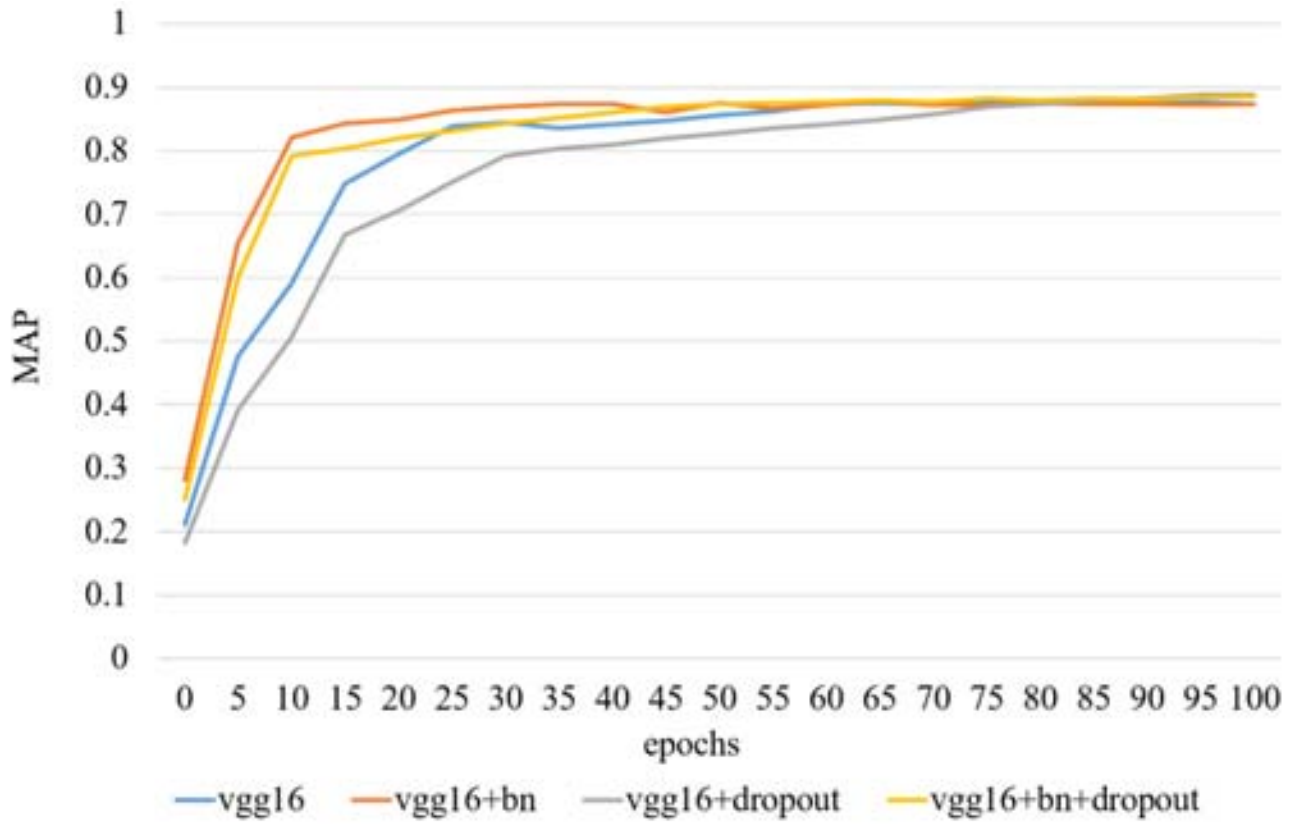

Fig.5 Influence of Batch Normalization and Dropout on the model training stage performance.

**Higher input image resolution improves model performance but is slower for training** In practice, zooming in the image could get more fine-grained information that helps radiologists in diagnosis. A previous work used M-CNN [21] and experimented on different image resolutions, finding that the input image resolution affects the model performance. We also tested a variety of input image resolutions, observing that higher resolution improves the model performance but decreases the inference speed.

TABLE II

THE EFFECT OF INPUT IMAGE RESOLUTION. HIGHER INPUT IMAGE RESOLUTION IMPROVES THE MODEL PERFORMANCE BUT DECREASES THE INFERENCE SPEED

| Image resolution | MAP | Speed(fps) |
|---|---|---|
| 256*256 | 0.627 | **56** |
| 320*320 | 0.776 | 36 |
| 512*512 | 0.852 | 26 |
| 640*640 | **0.882** | 20 |

TABLE III

MODEL PERFORMANCE OF RESIDUAL AND INCEPTION MODULE. RESIDUAL MODULE DOES NOT IMPROVE MODEL PERFORMANCE; INCEPTION MODULE DECREASES COMPUTATIONAL COMPLEXITY BUT ALSO DECREASES MAP

| Backbone model | MAP | Speed (fps) | Memory usage |
|---|---|---|---|
| VGG16+BN | **0.882** | 20 | ~4 GB |
| Resnet-18 | 0.528 | 34 | ~2 GB |
| Resnet-34 | 0.786 | 30 | ~2.9 GB |
| Resnet-50 | 0.827 | 26 | ~3.8 GB |
| Resnet-101 | 0.847 | 18 | ~5 GB |
| Inception-v3 | 0.826 | 36 | ~2.9 GB |

**Data augmentation is crucial** Since the training dataset only contains 7258 images with 7432 nodule candidates. Training a still image detector on such a small dataset makes model prone to overfitting. Data augmentation introduces noise to the model and enhances its robustness. We applied a series of data augmentation approaches and checked the performance:

-random flipping

-random cropping

-random adjustment of the hue, saturation, illumination and contrast.

All of the data augmentation was performed at the probability of 0.5. We obtained 9.6% MAP improvement. It is worth to note that adjusting hue, saturation, illumination and contrast makes model more robust to pneumonia patients.

**Residual and Inception module not works as expected** The residual module [12] uses the shortcut connection to reformulate neural network layers as the learning residual function with reference to the layer inputs, which enables ConvNets to go deeper to get extremely deep representations of image. The Inception module [20] splits the input into a few low-dimensional ones, transforms them by a set of filters (3×3, 5×5, etc.), which reduces the computational complexity and enables the network to extract features in a wider receptive field.

As shown in TABLE III, the model performance does not benefit from the residual module. Even the residual module helps us train an extremely deep ConvNet (101 layers), the performance cannot exceed the model performance without residual module (VGG16 and Inception-v3). The reason might be the residual module propagates too much noise to high-level features, which makes the model hard to learn efficient information of nodule.

The inception module works well in reducing computational complexity, but does not improve the model performance. This is not surprising because the inception module is designed to get information from multiple receptive fields, but for our task, the nodule size is almost fixed (from 3px to 40px, which is a small range compared with the image size).

### C. *MSP and MLGS*

For multi-slice propagator, building the temporal window *w* should not exceed the minimum lung nodule size, therefore we set *w=2* in experiment.

For motionless-guide suppressor, since we have already built a set of features to describe the nodule detected by still image detector and its motion information, we just need to adjust the hyperparameter of Random Forest. In this experiment, we set the number of trees in the forest at 100, and use 6 randomly picked features of all 40 features when splitting a tree node.

A feature importance analysis from Random Forest was presented in Fig.6. The most important feature is the still image score, which is not surprising because the nodule detection result is the composite of nodule information (size, density, edge characteristics, etc.), the second most important feature is the $d_{max}/n$ which indicates whether the nodule candidates suddenly appear and vanish. Also, the deviation feature is important.

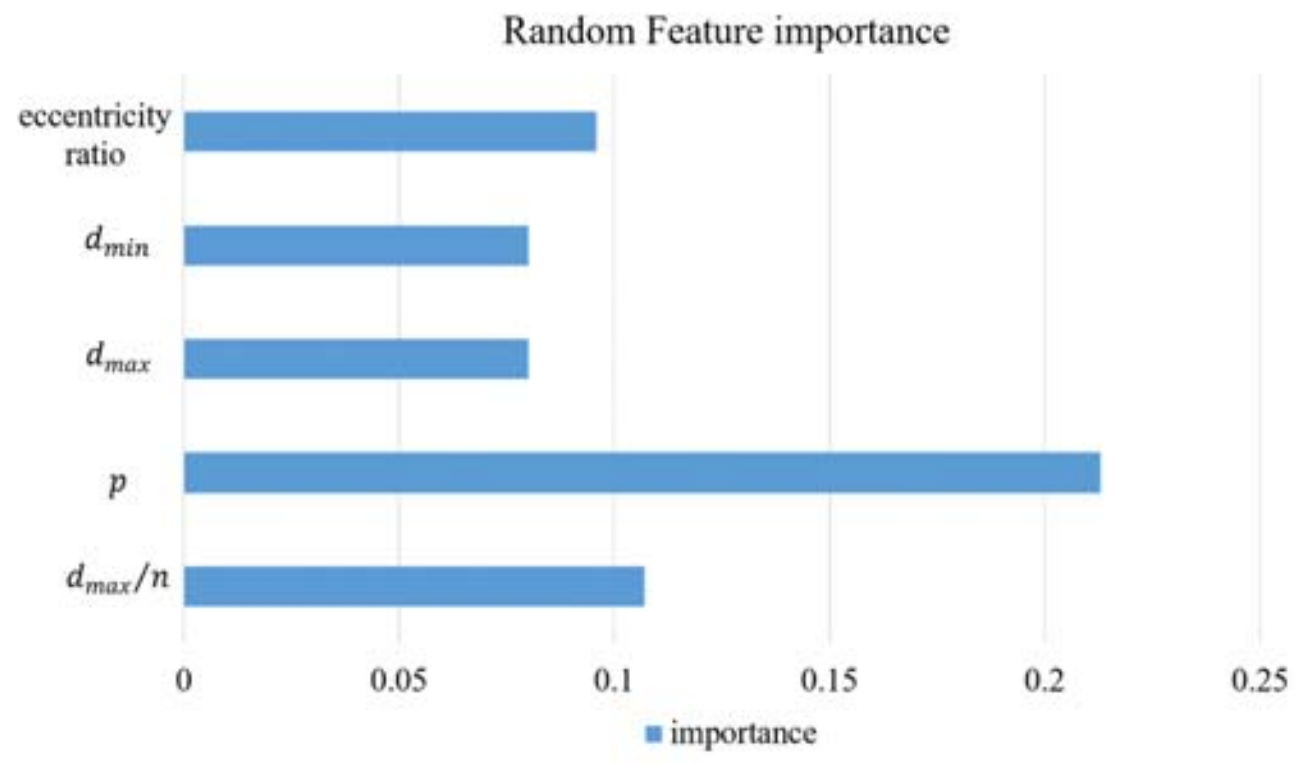

Fig.6 The top-5 important features obtained from Random Forest.

### *D. Result*

**Evaluation metric** According to the Luna16 standard, a predicted nodule candidate location was considered as a true positive if it was located within the radius of a true nodule center. We followed the LUNA16 guideline, evaluated the detection result by measuring the Free-Response Receiver Operating Characteristic (FROC) which calculates the average sensitivity over 7 false positive rates: 1/8, 1/4, 1/2, 1, 2, 4, 8 false positives per scan.

**Result of still image detector** We need to locate lung nodule candidates with high recall and low false positives, which is the key to utilize MSP and MLGS. We generated 5745 candidates with sensitivity of 97.63% (1158 out of 1186), outperforming the state-of-the-art lung nodule candidates generation algorithm in both recall and accuracy (see TABLE IV). We also compared the result with two baselines: 1) SSD without pyramid feature map, and 2) SSD without Focal Loss. The comparison between "MSSD" and "SSD no pyramid" indicates the importance of rich context high resolution features, while the comparison between "MSSD" and "SSD no focal" verifies the effectiveness of Focal Loss.

TABLE IV
PERFORMANCE OF STILL IMAGE DETECTOR (CANDIDATES GENERATOR)

| Method | Sensitivity | Candidates/per scan |
|---|---|---|
| ISICAD [7] | 0.856 | 335.9 |
| M5L [21] | 0.768 | 22.2 |
| Faster RCNN | 0.946 | 15 |
| Baseline no pyramid | 0.876 | 24.6 |
| Baseline no focal | 0.927 | 10.7 |
| MSSD | **0.976** | **6.46** |

TABLE V
COMPARISON OF OUR METHOD WITH OTHER CAD SYSTEMS IN FROC SCORE, SENSITIVITY AND SPEED

| Model | FROC | Sensitivity | Seconds/per scan |
|---|---|---|---|
| Multi-view | | 90.1 | - |
| 3D-CNN | 0.827 | 0.922 | - |
| Faster RCNN | 0.891 | 0.946 | 60 |
| MSSD+MSP+ MLGS | **0.892** | **0.953** | **20** |

'-' means this method is not an end-to-end solution.

**Result of the whole model** We compare performances of our final model with Multi-view model [3] which uses the approach of observing lung nodules in 9 different directions, Faster RCNN + C3D [1] which generates nodule candidates with Faster RCNN and reduces false positives with 3D ConvNets, and 3-D CNNs [2] which reduces false positives using multi-contexture information generated by 3D ConvNets. As shown in TABLE V, our model achieves FROC score of 0.892, sensitivity of 95.3% at 1 false positives per scan with high inference speed (within 20 seconds per patients, 2X faster than a Faster RCNN, 3X faster than radiologists). The MSP and MLGS reduce 65.6% of false positive detections (3013 out of 4587), which is an efficient false positive reduction algorithm and extremely fast (about 0.02 seconds for 300 candidates, 50x faster than Multi-view model). It is also worth to note that our model is not computationally intensive, which can run on a GPU with 4 GB memory.

The FROC curve is shown in Fig.7. Our model performance is robust, which ensures the model practicability in real-world lung nodule detection. Our model only generates 669 false positives, less than 1 false positives per scan, which satisfies clinical use standard.

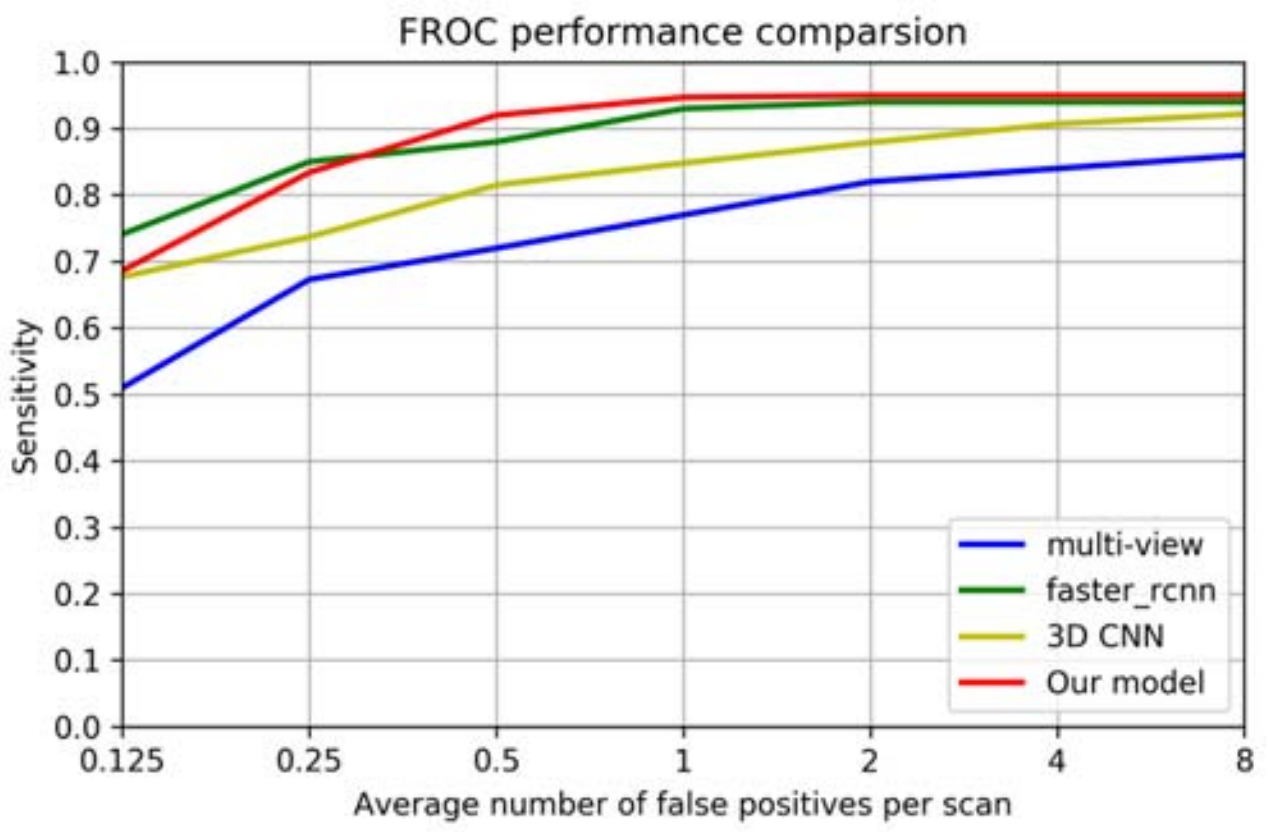

Fig.7 FROC curve. Our method outperforms state-of-the-art methods at the scan speed of more than 0.5 false positives per scan.

## IV. MODEL ANALYSIS AND CONCLUSION

### *A. Model analysis*

We visualized some qualified and unqualified detection results in Fig.8. The green and red rectangle respectively denote correct and incorrect detections. The correct detections show our model can detect nodules with various size, density, and edge characteristics. However, some false positives generate similar responses as the lung nodules and lead to incorrect detections. We list three kinds of typical false positives: 1) Tissues outside of lung may be very similar to the lung nodule such as the gastric area (Type-A in Fig.8). 2) The lung wall bulge generates similar responses with lung nodule invading into breast wall because we only observe the lung nodule in lung window. However, it is relatively easier to classify them in mediastinal window. This implies that we lost some information in data preprocessing step. A typical example was shown in Fig.8 Type-B. 3) The model may misclassify the nodule with wider blood vessel (Fig.8 Type-C) even with the motion information. This is because the blood vessel does not move significantly in adjacent slices.

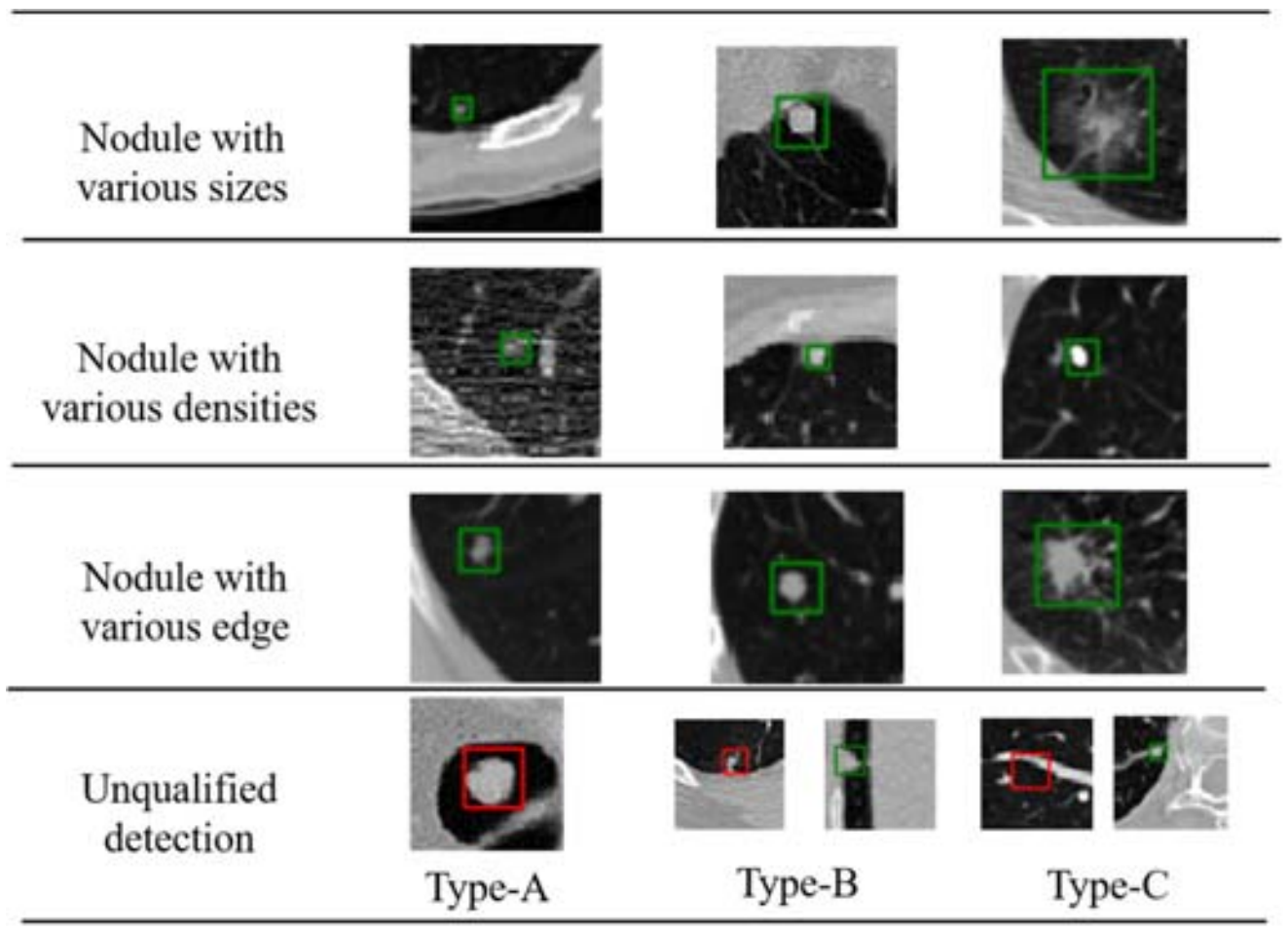

Fig.8 Quality detection result and some typical false positives. Green and red rectangles respectively indicate true and false positives. Our model is insensitive to nodule characteristics including size, density and edge. False positives are quite similar in appearance to true positives.

### B. *Discussion*

We propose a novel approach to detect pulmonary lung nodule in CT images by treating this task as an VID task and applying some technologies such as still image detector, motion information analysis, which lead to promising result with high detection result. We analyze the similarity and difference between CT images and video frames, improve the still image detector to enable it to learn rich context information such as size and density as well as keeping rich information of small lung nodule. We also propose two simple methods, MSP and MLSG, and utilize the motion information of lung nodules to reduce false negatives and false positives.

We can see that the CNN, being a dominant trend in natural image processing, pervade quickly in the medical image analysis community. However, there are still challenges when applying the deep learning technologies to medical image analysis: 1) some technologies that perform excellently on natural images may not work in medical image analysis. For example, we find that the residual model and inception module do not work as expected in detecting lung nodule in 2D images. 2) Labeled data with high quality are hard to acquire, which limits the deep learning usage in medical image analysis. 3) Training a reliable model from scratch is difficult, the prior knowledge of doctors is hard to integrate into the end-to-end deep learning structure.

The computation of deep ConvNet is intensive even with the high-end hardware and high-quality deep learning frame (e.g. MXNet [22]), which limits its clinical usage. With our light-weighted method, diagnosis of a patient needs 20 seconds with a GTX1080. How to incorporate radiologists' prior knowledge into deep ConvNet is still an open question.

## V. Acknowledgment

The authors would like to thank the LUNA16 challenge organizers for providing the dataset and evaluating our results.